\documentclass{article}

\usepackage[preprint]{neurips_2025}

\usepackage[utf8]{inputenc} 
\usepackage[T1]{fontenc}    
\usepackage{hyperref}       
\usepackage{url}            
\usepackage{booktabs}       
\usepackage{amsfonts}       
\usepackage{nicefrac}       
\usepackage{microtype}      
\usepackage{xcolor}         
\usepackage{graphicx, subcaption}

\usepackage{soul}
\usepackage{listings}
\usepackage{amsmath}

\title{Leveraging Compact Satellite Embeddings and Graph Neural Networks for Large-Scale Poverty Mapping}

\author{%
  Markus B.~Pettersson \\
  Chalmers University of Technology\\
  \texttt{markus.pettersson@chalmers.se} \\
  \And
  Adel Daoud \\
  Chalmers University of Technology \\
  Linköping University
}

\begin{document}

\maketitle

\begin{abstract}
  Accurate, fine-grained poverty maps remain scarce across much of the Global South. While Demographic and Health Surveys (DHS) provide high-quality socioeconomic data, their spatial coverage is limited and reported coordinates are randomly displaced for privacy, further reducing their quality. We propose a graph-based approach leveraging low-dimensional \textit{AlphaEarth} satellite embeddings to predict cluster-level wealth indices across Sub-Saharan Africa. By modeling spatial relations between surveyed and unlabeled locations, and by introducing a probabilistic ``fuzzy label'' loss to account for coordinate displacement, we improve the generalization of wealth predictions beyond existing surveys. Our experiments on 37 DHS datasets (2017–2023) show that incorporating graph structure slightly improves accuracy compared to ``image-only'' baselines, demonstrating the potential of compact EO embeddings for large-scale socioeconomic mapping.
\end{abstract}

\section{Introduction}
Accurate local-scale poverty estimates are crucial for targeting interventions \citep{daoudStatisticalModelingThree2023}, yet such data remain scarce across much of the developing world. The most reliable socioeconomic information in many countries still comes from household surveys such as the Demographic and Health Surveys (DHS) \citep{ICF_DHSWebsite}. Because these surveys cover only a limited set of locations, recent research has sought to predict local wealth or poverty indicators directly from satellite data \citep{jean2016combining, yeh2020using, pettersson2023time,daoud2023using}.

A common approach trains regression models to predict a survey-based wealth index, such as the International Wealth Index (IWI) used here \citep{smits2015international}, from remote-sensing imagery such as Landsat or Sentinel-2 \citep{pettersson2023time}. However, these methods often ignore spatial dependencies between nearby settlements. Neighboring communities tend to share infrastructure, markets, and environmental context, introducing autocorrelation that purely image-based models overlook. While including latitude and longitude as model inputs can partially capture this, such models fail to generalize to unseen countries and depend on absolute coordinates rather than spatial relationships.

A further complication arises from the way DHS anonymizes survey coordinates: each cluster’s location is randomly displaced by up to 2 km in urban and 10 km in rural areas \citep{BurgertEtAl2013_SAR7}. Consequently, the reported coordinates do not indicate the true village or neighborhood, introducing spatial uncertainty into all analyses \citep{kakooei2024increasing}.

To address both issues jointly, we propose representing survey and nearby settlement locations as nodes in a graph, where edges capture geographic proximity. Using compact, information-rich AlphaEarth embeddings derived from satellite imagery \citep{brown2025alphaearth}, we train neural and graph-based models that incorporate these relationships. We further introduce a ``fuzzy label'' loss function that explicitly accounts for the DHS coordinate perturbations by weighting candidate settlement locations according to their displacement likelihood. Together, these ideas allow wealth prediction models to generalize spatially while remaining computationally lightweight enough for continental-scale training.

\section{Data}
We used 37 DHS surveys conducted between 2017 and 2023 across 25 Sub-Saharan African countries, compromising approximately 16,600 clusters, roughly corresponding to villages in rural areas or neighborhoods in urban areas \cite{BurgertEtAl2013_SAR7}. For each cluster, we computed the mean International Wealth Index (IWI), a standardized measure of material wealth ranging from 0 to 100 \cite{smits2015international}.

To preserve respondent privacy, DHS displaces cluster coordinates randomly: urban clusters are perturbed by up to 2 km, and rural clusters by up to 10 km. This introduces spatial noise, making the reported coordinates approximate rather than exact; a key motivation for our ``fuzzy label'' approach. A more detailed discussion on this can be found in Appendix \ref{sec:displacement_mechanism}.

As a reference set of potential settlement locations in the given countries, we extracted >167,000 points labeled as ``Populated places'' from the GeoNames database \citep{geonames}. While this dataset is incomplete and is sure to contain selection bias (i.e., places with better connectivity reports more than those will worse), it provides a practical set of candidate locations for modeling both surveyed and surveyed villages.

For every location, whether from DHS or GeoNames, we obtain a 64-dimensional ALphaEarth embedding from Google Earth Engine, using a 500 m buffer around each coordinate and retrieving the mean pixel value \citep{brown2025alphaearth, gorelick2017google}. These embeddings provide compact, informative representations of local land use and environmental context, which serve as the primary input for all our predictive models.

The smaller footprint allows us to work with continental scale data in memory, opening the door for utilizing powerful graph architectures. As an example, the corresponding dataset using Sentinel-2 data would require 51.5 GB, but with AlphaEarth, this becomes 45.6 MB, easily small enough to fit on any GPU \citep{ESA_Sentinel2UserHandbook}.

\section{Method}

\begin{figure}
 \begin{subfigure}{0.16\textwidth}
     \includegraphics[width=\textwidth]{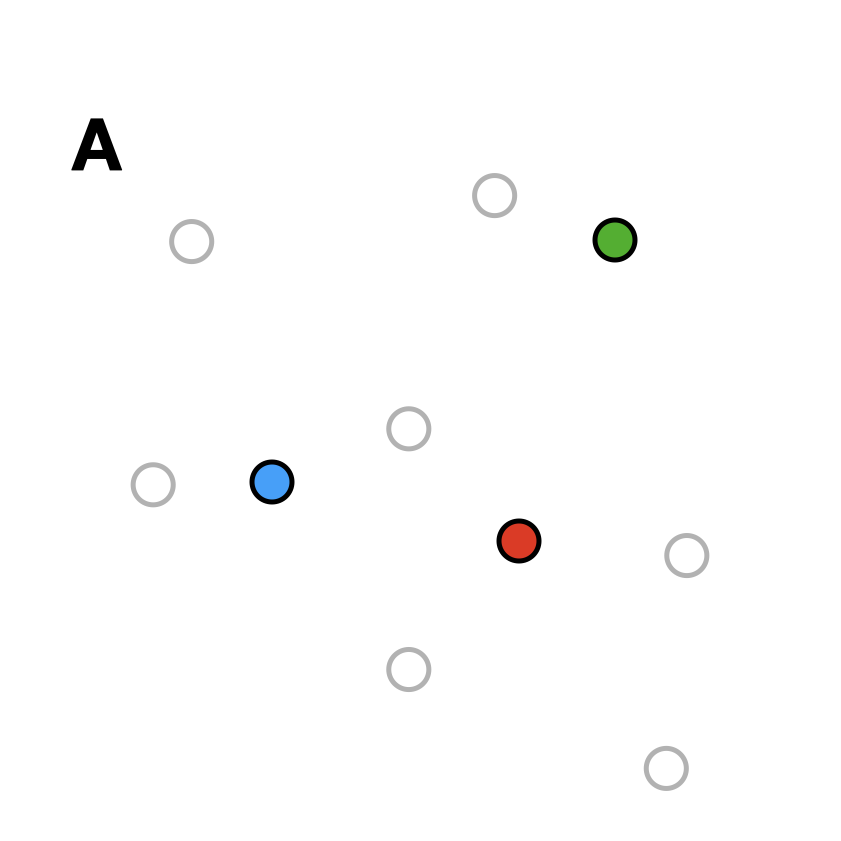}
 \end{subfigure}
 \hfill
 \begin{subfigure}{0.16\textwidth}
     \includegraphics[width=\textwidth]{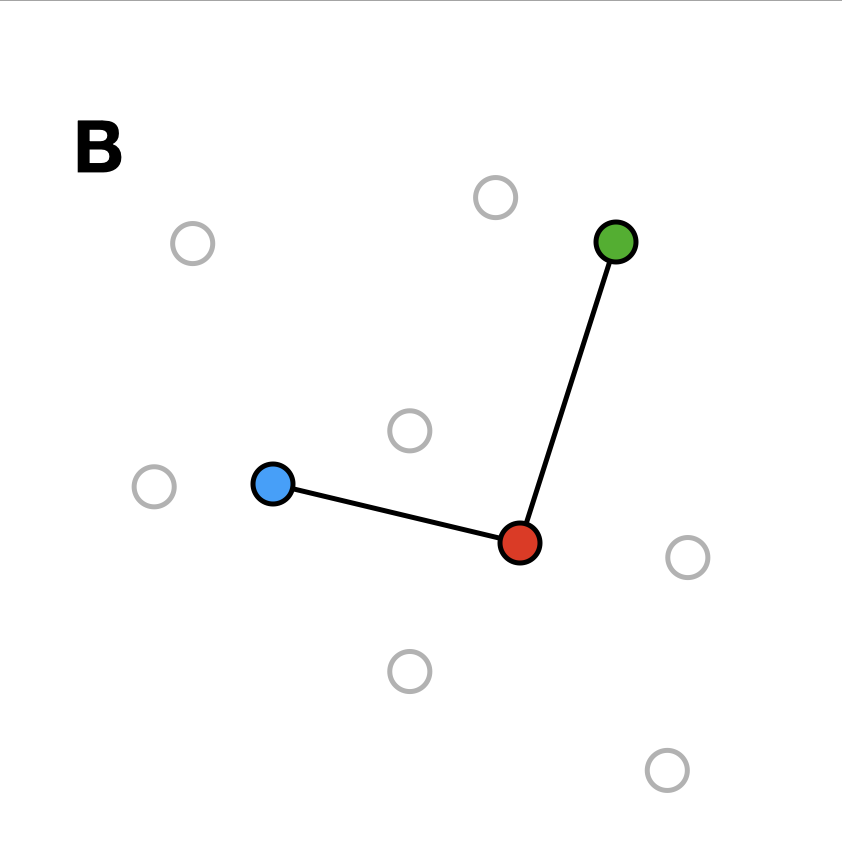}
 \end{subfigure}
 \hfill
 \begin{subfigure}{0.16\textwidth}
     \includegraphics[width=\textwidth]{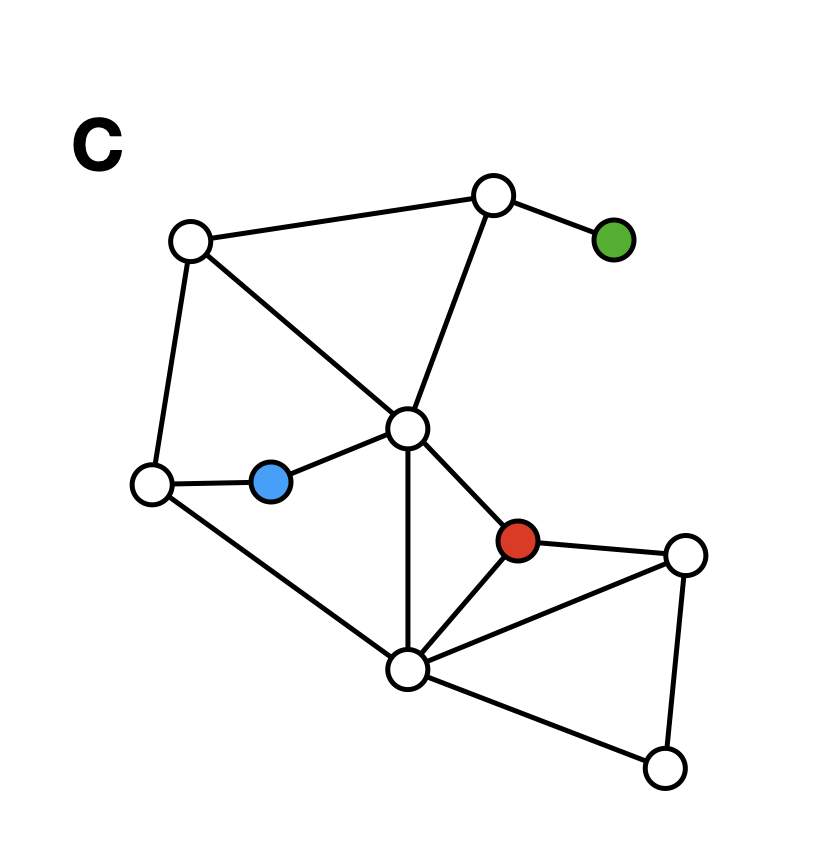}
 \end{subfigure}
 \hfill
 \begin{subfigure}{0.16\textwidth}
     \includegraphics[width=\textwidth]{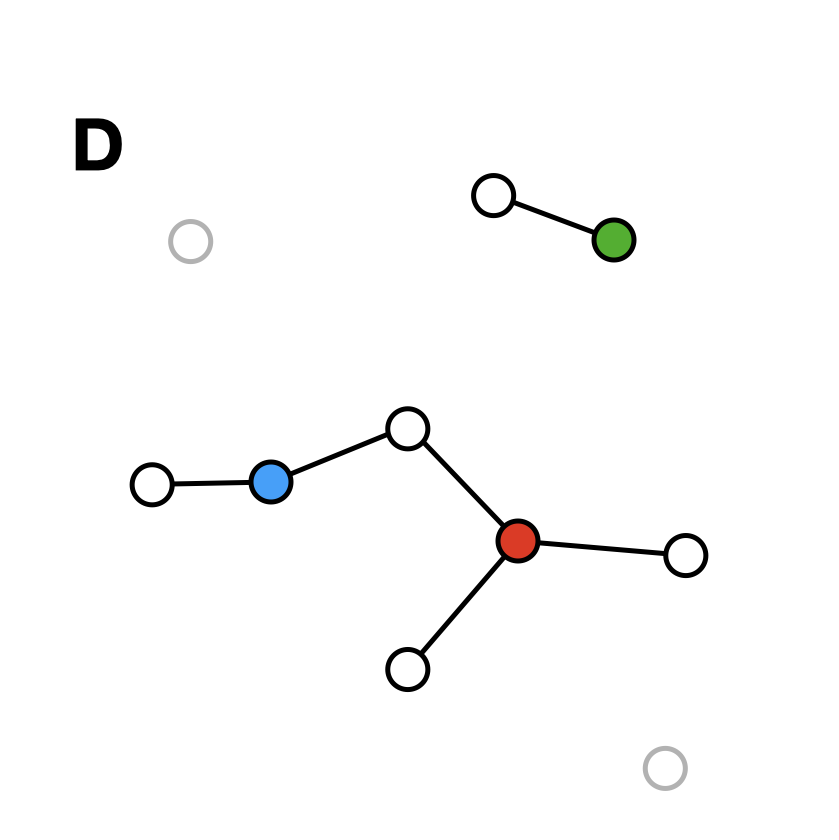}
 \end{subfigure}
 \hfill
 \begin{subfigure}{0.16\textwidth}
     \includegraphics[width=\textwidth]{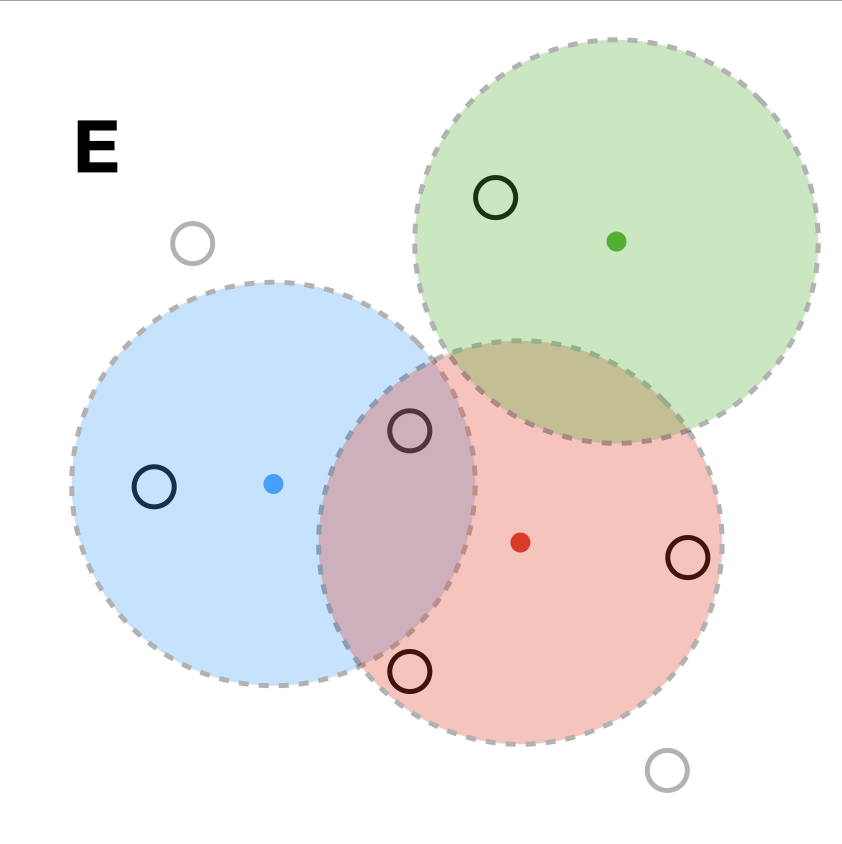}
 \end{subfigure}
 \hfill
 \begin{subfigure}{0.16\textwidth}
     \includegraphics[width=\textwidth]{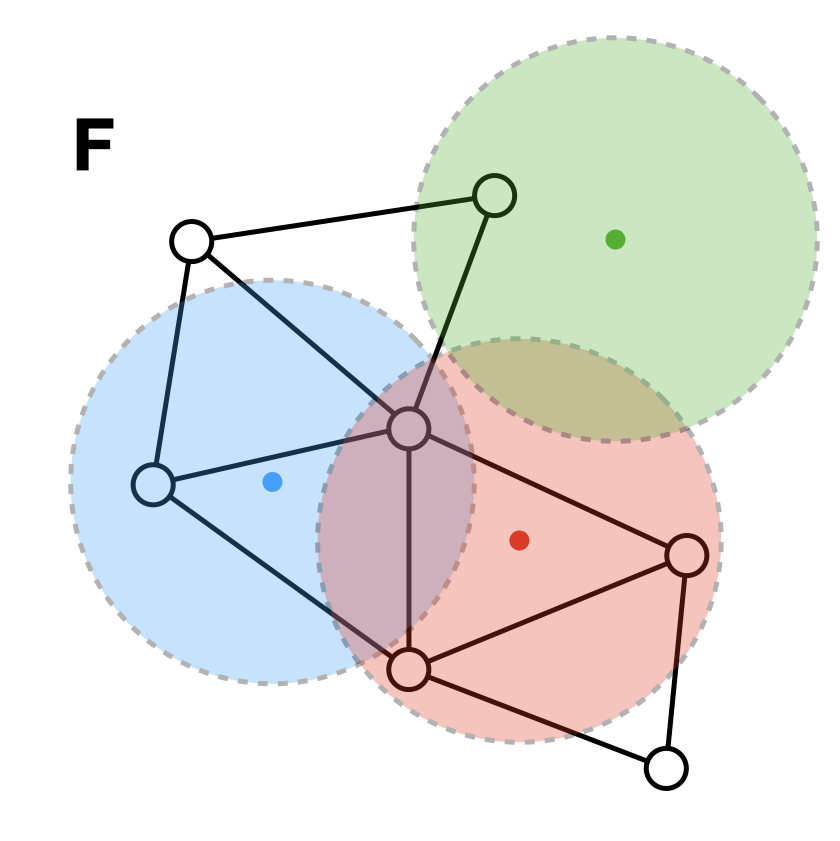}
 \end{subfigure}

 \caption{Structures of the six methods evaluated. In this figure, The colored nodes are different survey locations from the DHS while the the white are settlement locations from GeoNames. Both of these have corresponding AlphaEarth embeddings $x_i$. The grayed out GeoNames locations in A, B, D and E signify that they are not included in the model training. For E and F, where the embeddings of the survey clusters are not included, the larger circle highlights the possible displacement area which assigns fuzzy labels to nearby GeoNames nodes.}
 \label{fig:variants}
\end{figure}

We evaluate six modeling approaches that combine three design dimensions: (1) whether spatial relationships are modeled using a graph; (2) whether auxiliary unlabeled locations are included; and (3) whether fuzzy labels are used to account for coordinate displacement. Figure \ref{fig:variants} summarizes these variants.

\paragraph{(A) Baseline: DHS Points.}
A simple multilayer perceptron predicts cluster-level IWI directly from the AlphaEarth embedding at the reported survey coordinate. This represents the standard ``image-only'' approach.

\paragraph{(B) DHS graph.}
We connect DHS clusters within each survey into a graph, drawing edges between locations less than 100 km apart, weighted inversely by distance. A graph convolutional network (GCN) then propagates information across nearby clusters \citep{kipf2016semi}.

\paragraph{(C) Full graph (DHS + GeoNames).}
We expand the graph by adding unlabeled settlement nodes from the GeoNames database, linking each to its eight nearest neighbors (within 100 km). This allows the model to reason about relationships between surveyed and unsurveyed places.

\paragraph{(D) Ego graphs.}
For each DHS cluster, we construct a local subgraph connecting it to all nearby GeoNames settlements within its possible displacement radius (2-10 km). Edge weights represent the probability that each candidate is the true location given the DHS displacement mechanism.

To incorporate the uncertainty in cluster coordinates, we define a ``fuzzy label loss function'':

$$
\mathcal{L} = \frac{1}{N_l}\sum_{j=1}^{N_l}\sum_{i=1}^{N_s} P_{ji},(f_\theta(x_i)-y_j)^2
$$
where ($y_j$) is the wealth label for survey cluster ($j$), ($x_i$) is the embedding for candidate location ($i$), and ($P_{ji}$) is the normalized probability that ($i$) is the true position of cluster ($j$). This formulation effectively spreads each label across its plausible true locations.

\paragraph{(E) Points With Fuzzy labels.}
A non-graph version using the fuzzy loss, predicting wealth from embeddings of candidate locations only. The <3 \% of reported survey coordinates with no candidate location within its possible displacement radius (due to missingness in GeoNames) gets included in the set as a ``phantom location''.

\paragraph{(F) Graph With Fuzzy labels.}
A full GCN trained with the fuzzy-label objective, allowing message passing between both fuzzy-labeled, unlabeled, and phantom nodes.
\subsection{Model and training}
For all methods, we used small three-layer neural networks or equivalent GCNs with matched parameter counts for graphs. Methods A-D were trained with Mean-squared error loss (MSE), while E and F used the Fuzzy loss descibed above. Cross-validation was conducted across surveys to test generalization to unseen national contexts. More details on models and the cross validation can be found in Appendix \ref{sec:model_details}.

\section{Results}

We evaluate the performance of the methods on held-out cross-validation folds and the results can be found in Table \ref{table:results}.

\begin{table}[h]
\caption{Results for the different method (mean $\pm$ standard deviation across CV folds). Lower MAE and higher $R^2$ indicate better performance.}
\begin{tabular}{|l|ccc|c|c|}
\hline
\textbf{Method}     & \multicolumn{1}{c|}{GeoNames} & \multicolumn{1}{c|}{Graph} & Fuzzy labels              & MAE ↓                      & $R^2$ ↑                    \\ \hline
(A) DHS Points      &                               &                            &                           & $8.657 \pm 0.343$          & $0.546 \pm 0.047$          \\
(B) DHS Graph       &                               & \checkmark  &                           & $8.682 \pm 0.367$          & $0.546 \pm 0.040$          \\
(C) Full Graph      & \checkmark     & \checkmark  &                           & $8.588 \pm 0.472$          & $0.546 \pm 0.043$          \\
(D) Ego Graphs      & \checkmark     & \checkmark  &                           & $\mathbf{8.354 \pm 0.436}$ & $\mathbf{0.569 \pm 0.047}$ \\
(E) Points w. Fuzzy & \checkmark     &                            & \checkmark & $9.217 \pm 0.539$          & $0.465 \pm 0.075$          \\
(F) Graph w. Fuzzy  & \checkmark     & \checkmark  & \checkmark & $9.665 \pm 0.322$          & $0.406 \pm 0.084$          \\ \hline
\end{tabular}
\label{table:results}
\end{table}

Models incorporating spatial structure and unlabeled settlements (C–D) modestly outperform the image-only baseline (A), suggesting that nearby settlements contain complementary information not captured by single-site embeddings. The Ego Graphs model (D) performs best overall, indicating that constraining message passing to the plausible displacement radius helps leverage local spatial context without introducing noise from distant or unrelated nodes.

In contrast, models trained with fuzzy labels (E–F) perform notably worse. This suggests that while accounting for coordinate uncertainty is conceptually appealing, the reported displaced coordinates may still provide more reliable supervision than the set of nearby candidate locations. Possible reasons for this are discussed in the following section.

\begin{figure}
    \centering
    \includegraphics[width=\linewidth]{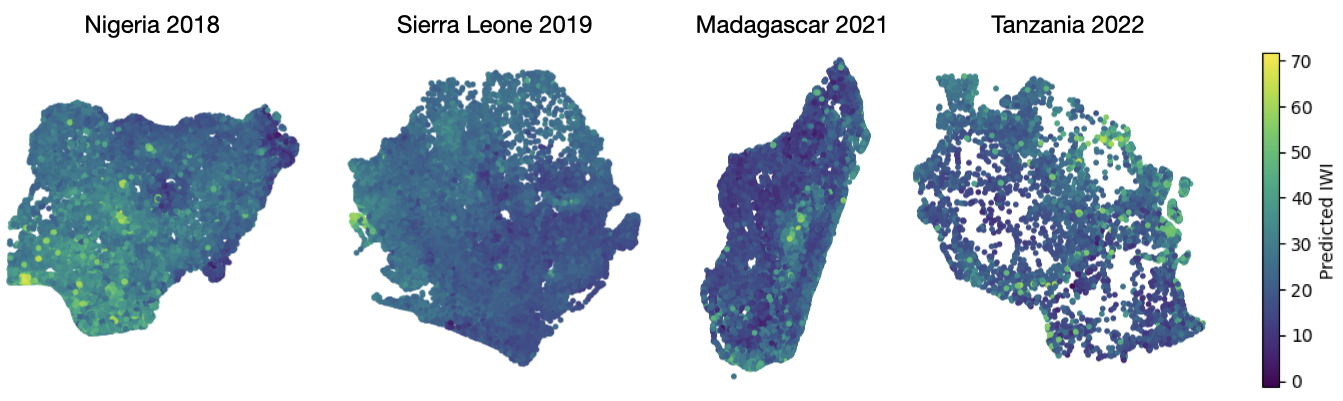}
    \caption{Predicted cluster-level IWI across GeoNames settlements in selected countries. Coverage and spatial density vary by country.}
    \label{fig:prediction_maps}
\end{figure}

Figure \ref{fig:prediction_maps} illustrates model predictions across all GeoNames settlements for a subset of countries. The resulting maps exhibit plausible spatial patterns, with smoother regional trends and higher predicted wealth near urban centers, broadly consistent with previous findings \citep{worldbank2009}.

\section{Discussion and Outlook}
The poor performance of fuzzy-label variants highlights the difficulty of learning under spatial uncertainty. Although the DHS coordinate perturbation motivates such and approach, the GeoNames dataset is incomplete and likely biased toward larger or more accessible settlements. In addition, our current method does not account for the prior of the DHS displacement mechanism: in order to make a representative survey, clusters are more likely to be located in settlements with higher population. Our current probabilistic weighting treats any two settlements with the same distance to the reported coordinate as equally likely, underestimating this prior knowledge. Future work could address these issues by incorporating ancillary data such as population density when defining candidate locations and label weights \citep{kakooei2024increasing}. Otherwise, such biases could lead to e.g. misallocation of aid resources due to biased mapping.

Beyond methodological refinements, our experiments demonstrate the practicality of large scale graph learning with modern Earth observation embeddings. The compactness of representations such as AlphaEarth enables training on national or continental graphs with relatively minimal computational cost, making spatially aware poverty mapping feasible even on modest hardware. In the future, other embeddings such as ESA's Terramind or Clay could be included for comparison \citep{jakubik2025terramind, ClayFoundationModel}.

Moving forward, several directions appear promising. First, refining the representation of spatial relationships could yield substantial gains. Rather than connecting nodes purely by Euclidean distance, edges could be weighted by estimated travel time derived from road networks or terrain models, better capturing real-world accessibility and socioeconomic interaction. Second, obtaining and incorporating the census enumeration areas or sampling frames actually used by DHS would provide a much more accurate set of candidate locations, replacing the incomplete GeoNames database. This, in combination with population weighting, would allow fuzzy-label models to operate on realistic spatial priors tied to how survey clusters are selected in practice. Finally, future work could explore attention-based or hierarchical graph neural networks and population- or time-weighted embeddings to capture multi-scale and temporal socioeconomic dynamics.

In summary, our study shows that high-quality, low-dimensional EO representations make it possible to train graph-based models at scale. We also show that encoding spatial relationships, especially within the known uncertainty bounds of survey locations, offers a tangible path toward more reliable, fine-grained poverty estimation in data-scarce regions.

\bibliographystyle{plainnat}

\begin{thebibliography}{20}
\providecommand{\natexlab}[1]{#1}
\providecommand{\url}[1]{\texttt{#1}}
\expandafter\ifx\csname urlstyle\endcsname\relax
  \providecommand{\doi}[1]{doi: #1}\else
  \providecommand{\doi}{doi: \begingroup \urlstyle{rm}\Url}\fi

\bibitem[Bank(2009)]{worldbank2009}
World Bank.
\newblock World development report 2009: Reshaping economic geography.
\newblock Technical report, World Bank, 2009.
\newblock URL \url{http://hdl.handle.net/10986/5991}.
\newblock License: CC BY 3.0 IGO.

\bibitem[Brown et~al.(2025)Brown, Kazmierski, Pasquarella, Rucklidge, Samsikova, Zhang, Shelhamer, Lahera, Wiles, Ilyushchenko, et~al.]{brown2025alphaearth}
Christopher~F Brown, Michal~R Kazmierski, Valerie~J Pasquarella, William~J Rucklidge, Masha Samsikova, Chenhui Zhang, Evan Shelhamer, Estefania Lahera, Olivia Wiles, Simon Ilyushchenko, et~al.
\newblock Alphaearth foundations: An embedding field model for accurate and efficient global mapping from sparse label data.
\newblock \emph{arXiv preprint arXiv:2507.22291}, 2025.

\bibitem[Burgert et~al.(2013)Burgert, Colston, Roy, and Zachary]{BurgertEtAl2013_SAR7}
Clara~R. Burgert, Josh Colston, Thea Roy, and Blake Zachary.
\newblock Geographic displacement procedure and georeferenced data release policy for the demographic and health surveys.
\newblock Dhs spatial analysis reports no.\ 7, The DHS Program / ICF International, Calverton, Maryland, USA, September 2013.
\newblock URL \url{https://www.dhsprogram.com/publications/publication-SAR7-Spatial-Analysis-Reports.cfm}.
\newblock Publication ID: SAR7. Available from The DHS Program website.

\bibitem[{Clay Foundation}(2025)]{ClayFoundationModel}
{Clay Foundation}.
\newblock Clay foundation model, 2025.
\newblock URL \url{https://clay-foundation.github.io/model/index.html}.
\newblock Accessed: 2025-10-21.

\bibitem[Daoud and Dubhashi(2023)]{daoudStatisticalModelingThree2023}
Adel Daoud and Devdatt Dubhashi.
\newblock Statistical {{Modeling}}: {{The Three Cultures}}.
\newblock 5\penalty0 (1), 2023.
\newblock ISSN 2644-2353, 2688-8513.
\newblock \doi{10.1162/99608f92.89f6fe66}.
\newblock URL \url{https://hdsr.mitpress.mit.edu/pub/uo4hjcx6/release/1}.

\bibitem[Daoud et~al.(2023)Daoud, Jord{\'a}n, Sharma, Johansson, Dubhashi, Paul, and Banerjee]{daoud2023using}
Adel Daoud, Felipe Jord{\'a}n, Makkunda Sharma, Fredrik Johansson, Devdatt Dubhashi, Sourabh Paul, and Subhashis Banerjee.
\newblock Using satellite images and deep learning to measure health and living standards in india.
\newblock \emph{Social Indicators Research}, 167\penalty0 (1):\penalty0 475--505, 2023.

\bibitem[{European Space Agency}(2015)]{ESA_Sentinel2UserHandbook}
{European Space Agency}.
\newblock \emph{Sentinel-2 User Handbook}.
\newblock European Space Agency, 2015.
\newblock URL \url{https://sentinels.copernicus.eu/documents/247904/685211/Sentinel-2_User_Handbook}.
\newblock Accessed October 20, 2025.

\bibitem[Fey and Lenssen(2019)]{fey2019fast}
Matthias Fey and Jan~Eric Lenssen.
\newblock Fast graph representation learning with pytorch geometric.
\newblock \emph{arXiv preprint arXiv:1903.02428}, 2019.

\bibitem[Gorelick et~al.(2017)Gorelick, Hancher, Dixon, Ilyushchenko, Thau, and Moore]{gorelick2017google}
Noel Gorelick, Matt Hancher, Mike Dixon, Simon Ilyushchenko, David Thau, and Rebecca Moore.
\newblock Google earth engine: Planetary-scale geospatial analysis for everyone.
\newblock \emph{Remote Sensing of Environment}, 2017.
\newblock \doi{10.1016/j.rse.2017.06.031}.
\newblock URL \url{https://doi.org/10.1016/j.rse.2017.06.031}.

\bibitem[Hamilton et~al.(2017)Hamilton, Ying, and Leskovec]{hamilton2017inductive}
Will Hamilton, Zhitao Ying, and Jure Leskovec.
\newblock Inductive representation learning on large graphs.
\newblock \emph{Advances in neural information processing systems}, 30, 2017.

\bibitem[{ICF}(2025)]{ICF_DHSWebsite}
{ICF}.
\newblock {The DHS Program}.
\newblock \url{https://www.dhsprogram.com}, 2025.
\newblock Funded by USAID. Accessed October 20, 2025.

\bibitem[Jakubik et~al.(2025)Jakubik, Yang, Blumenstiel, Scheurer, Sedona, Maurogiovanni, Bosmans, Dionelis, Marsocci, Kopp, et~al.]{jakubik2025terramind}
Johannes Jakubik, Felix Yang, Benedikt Blumenstiel, Erik Scheurer, Rocco Sedona, Stefano Maurogiovanni, Jente Bosmans, Nikolaos Dionelis, Valerio Marsocci, Niklas Kopp, et~al.
\newblock Terramind: Large-scale generative multimodality for earth observation.
\newblock \emph{arXiv preprint arXiv:2504.11171}, 2025.

\bibitem[Jean et~al.(2016)Jean, Burke, Xie, Alampay~Davis, Lobell, and Ermon]{jean2016combining}
Neal Jean, Marshall Burke, Michael Xie, W~Matthew Alampay~Davis, David~B Lobell, and Stefano Ermon.
\newblock Combining satellite imagery and machine learning to predict poverty.
\newblock \emph{Science}, 353\penalty0 (6301):\penalty0 790--794, 2016.

\bibitem[Kakooei and Daoud(2024)]{kakooei2024increasing}
Mohammad Kakooei and Adel Daoud.
\newblock Increasing the confidence of predictive uncertainty: earth observations and deep learning for poverty estimation.
\newblock \emph{IEEE Transactions on Geoscience and Remote Sensing}, 62:\penalty0 1--13, 2024.

\bibitem[Kipf(2016)]{kipf2016semi}
TN~Kipf.
\newblock Semi-supervised classification with graph convolutional networks.
\newblock \emph{arXiv preprint arXiv:1609.02907}, 2016.

\bibitem[Loshchilov and Hutter(2017)]{loshchilov2017decoupled}
Ilya Loshchilov and Frank Hutter.
\newblock Decoupled weight decay regularization.
\newblock \emph{arXiv preprint arXiv:1711.05101}, 2017.

\bibitem[Pettersson et~al.(2023)Pettersson, Kakooei, Ortheden, Johansson, and Daoud]{pettersson2023time}
Markus~B Pettersson, Mohammad Kakooei, Julia Ortheden, Fredrik~D Johansson, and Adel Daoud.
\newblock Time series of satellite imagery improve deep learning estimates of neighborhood-level poverty in africa.
\newblock In \emph{IJCAI}, pages 6165--6173, 2023.

\bibitem[Smits and Steendijk(2015)]{smits2015international}
Jeroen Smits and Roel Steendijk.
\newblock The international wealth index (iwi).
\newblock \emph{Social indicators research}, 122\penalty0 (1):\penalty0 65--85, 2015.

\bibitem[Wick(2005)]{geonames}
Marc Wick.
\newblock Geonames geographical database, 2005.
\newblock URL \url{https://www.geonames.org}.
\newblock Accessed: 2025-10-11.

\bibitem[Yeh et~al.(2020)Yeh, Perez, Driscoll, Azzari, Tang, Lobell, Ermon, and Burke]{yeh2020using}
Christopher Yeh, Anthony Perez, Anne Driscoll, George Azzari, Zhongyi Tang, David Lobell, Stefano Ermon, and Marshall Burke.
\newblock Using publicly available satellite imagery and deep learning to understand economic well-being in africa.
\newblock \emph{Nature communications}, 11\penalty0 (1):\penalty0 2583, 2020.

\end{thebibliography}

\newpage
\appendix

\section{Displacement mechanism}
\label{sec:displacement_mechanism}
The DHS cluster displacement procedure, described in detail by \citet{smits2015international}, anonymizes household survey coordinates through a two-step randomization process. First, a random direction is drawn uniformly from $\left[0, 2\pi\right)$. Second, a displacement distance is sampled according to cluster type, which is reported for all clusters in the survey. For urban clusters, the distance is drawn uniformly from 0–2 km. For rural clusters, 99\% are displaced by a random distance uniformly distributed between 0–5 km, and the remaining 1\% between 0–10 km. The reported cluster coordinate is then obtained by displacing the true location by the sampled distance along the sampled direction. This randomized offset ensures respondent privacy while preserving approximate spatial structure for analytical use.

Let $d$ denote the distance between the true and reported coordinates. The likelihood of observing the reported coordinate given the true location will thus be

$$
L(d) = 
\begin{cases}
    \frac{1}{\pi 2^2} & d \leq 2 \\
    0 & d > 2
\end{cases}
$$

for urban clusters and

$$
L(d) = 
\begin{cases}
    \frac{0.99}{\pi 5^2} + \frac{0.01}{\pi 10^2} & d \leq 5 \\
    \frac{0.01}{\pi 10^2} & 5 < d \leq 10 \\
    0 & d > 10
\end{cases}
$$

for rural clusters.

\section{Model details}
\label{sec:model_details}

As was described in the text, we select one of two basic model architectures for all methods, depending on whether it uses a graph structure or not. The models were implemented in PyTorch, using the PyTorch Geometric package for GCN Convolution \citep{fey2019fast}. These two architectures are included below:

\begin{lstlisting}[language=Python, caption={Simple Feedforward Neural Network used for methods (A) and (E).}]
class SimpleNN(torch.nn.Module):
    def __init__(self, input_dim, hidden_dim, output_dim):
        super().__init__()
        self.fc1 = torch.nn.Linear(input_dim, hidden_dim)
        self.fc2 = torch.nn.Linear(hidden_dim, hidden_dim)
        self.fc3 = torch.nn.Linear(hidden_dim, output_dim)

    def forward(self, x, edge_index, edge_attr=None, batch=None):
        # edge_index, edge_attr, and batch are unused in this model
        x = F.relu(self.fc1(x))
        x = F.relu(self.fc2(x))
        x = self.fc3(x)
        return x
\end{lstlisting}

\begin{lstlisting}[language=Python, caption={Simple Graph Neural Network  used for methods (B), (C), (D) and (F).}]
class SimpleGNN(torch.nn.Module):
    def __init__(self, input_dim, hidden_dim, output_dim):
        super().__init__()
        self.conv1 = GCNConv(input_dim, hidden_dim)
        self.conv2 = GCNConv(hidden_dim, hidden_dim)
        self.lin = torch.nn.Linear(hidden_dim, output_dim)

    def forward(self, x, edge_index, edge_attr=None, batch=None):
        # edge_attr here is inverse distance
        edge_weight = edge_attr.view(-1)
        x = self.conv1(x, edge_index, edge_weight=edge_weight)
        x = F.relu(x)
        x = self.conv2(x, edge_index, edge_weight=edge_weight)
        x = F.relu(x)

        x = self.lin(x)
        return x
\end{lstlisting}

For all methods, we used an input dimension of 64 (from the AlphaEarth embeddings), a hidden dimension of 128 and an output dimension of 1 (from the scalar IWI). The two models have the same number of parameters, to keep comparisons fair. For each method, we trained the corresponding model with Mean-Squared Error Loss (except for method E and F which use the Fuzzy loss described in the paper) and AdamW for 500 epochs, using checkpointing for the lowest validation loss \citep{loshchilov2017decoupled}. We did a small grid search for learning rate on the cross-validation folds, but otherwise no hyperparameter tuning.

Although the smaller footprint of the AlphaEarth embeddings allows us to fit models on the full graphs in parallel, we found better performance when subsampling neighborhoods of the graph as described by \citet{hamilton2017inductive}. For each labeled node, a subgraph is created by randomly sampling four of it's first degree neighbors and 16 of it's second degree neighbors. This improvement is likely due to the added stochasticity of mini-batching.

All the models were trained on a single A100 GPU and they all converged within 15 minutes, with the majority of the computational time for the graph models being devoted to sampling subgraphs which is done on the CPU.

\section{Cross-Validation Folds}

\begin{table}[ht]
\centering
\caption{Cross-validation splits for the different surveys.}
\begin{tabular}{|c|p{12cm}|}
\hline
\textbf{Splits} & \textbf{Surveys} \\
\hline
A & Kenya 2022 Standard DHS, Rwanda 2019-20 Standard DHS, Guinea 2018 Standard DHS, Ethiopia 2019 Interim DHS, Senegal 2018 Continuous DHS, Guinea 2021 MIS \\
\hline
B & Nigeria 2018 Standard DHS, Zambia 2018 Standard DHS, Cameroon 2018 Standard DHS, Liberia 2019-20 Standard DHS, Burkina Faso 2017-18 MIS, Mali 2021 MIS, Togo 2017 MIS \\
\hline
C & Mauritania 2019-21 Standard DHS, Sierra Leone 2019 Standard DHS, Burkina Faso 2021 Standard DHS, Mali 2018 Standard DHS, Uganda 2018-19 MIS, Senegal 2019 Continuous DHS, Niger 2021 MIS \\
\hline
D & Madagascar 2021 Standard DHS, Nigeria 2021 MIS, Benin 2017-18 Standard DHS, Cameroon 2022 MIS, Senegal 2017 Continuous DHS, Kenya 2020 MIS, Senegal 2020-21 MIS, Ghana 2019 MIS \\
\hline
E & Tanzania 2022 Standard DHS, Ghana 2022 Standard DHS, Cote d'Ivoire 2021 Standard DHS, Tanzania 2017 MIS, Gabon 2019-21 Standard DHS, Gambia 2019-20 Standard DHS, Mozambique 2018 MIS, Liberia 2022 MIS, Malawi 2017 MIS \\
\hline
\end{tabular}
\end{table}

\begin{table}[h]
    \centering
    \caption{Cross-validation data splits for each fold}
    \begin{tabular}{|c|c|c|}
    \hline \textbf{Train} & \textbf{Val} & \textbf{Test} \\
    \hline CDE & B & A \\
    ADE & C & B \\
    ABE & D & C \\
    ABC & E & D \\
    BCD & A & E \\
    \hline
    \end{tabular}
\end{table}

\end{document}